\documentclass[pdflatex,sn-mathphys]{sn-jnl}

\jyear{2021}%

\theoremstyle{thmstyleone}%
\theoremstyle{thmstyletwo}%

\theoremstyle{thmstylethree}%

\usepackage{graphicx}
\usepackage{xcolor}
\usepackage{hyperref}

\usepackage{pifont}

\usepackage{soul,color}
\soulregister\cite7
\soulregister\ref7

\usepackage{xcolor}
\newcommand{\change}[1]{\textcolor{black}{#1}} 

\begin{document}

\title[Article Title]{Task-Aware Asynchronous \change{Multi-Task Model} with  Class Incremental Contrastive Learning for Surgical Scene Understanding}

\author{Lalithkumar Seenivasan$^{1}$, Mobarakol Islam$^{2}$, Mengya Xu$^{1}$, Chwee Ming  Lim$^{3}$ 
        and Hongliang Ren$^{1,4}$\footnote{ Corresponding to Hongliang Ren hlren@ieee.org.\\
        $^{1}$Dept. of Biomedical Engineering, National University of Singapore, Singapore.\\
        $^{2}$Dept. of Computing, Imperial College London, United Kingdom. \\
        $^{3}$Head \& Neck Surgery, Singapore General Hospital, Singapore. \\
        $^{4}$Dept. of Electrical Engineering and Shun Hing Institute of Advanced Engineering, The Chinese University of Hong Kong.}

}

\abstract{
\textbf{Purpose:}
Surgery scene understanding with tool-tissue interaction recognition and automatic report generation can play an important role in intra-operative guidance, decision-making and postoperative analysis in robotic surgery. However, domain shifts between different \change{surgeries} with inter and intra-patient variation and novel instruments' appearance degrade \change{the performance of model prediction}. Moreover, it requires output from multiple models, which can be computationally expensive and affect real-time performance. 

\textbf{Methodology:} 
A multi-task learning (MTL) model is proposed for surgical report generation and tool-tissue interaction prediction that deals with domain shift problems. The model forms of shared feature extractor, mesh-transformer branch for captioning and graph attention branch for tool-tissue interaction prediction. The shared feature extractor employs class incremental contrastive learning (CICL) to tackle intensity shift and novel class appearance in the target domain. We design Laplacian of Gaussian (LoG) based curriculum learning into both shared and task-specific branches to enhance model learning. We incorporate a task-aware asynchronous MTL optimization technique to fine-tune the shared weights and converge both tasks \change{optimally}.

\textbf{Results:} 
The proposed MTL model trained using task-aware optimization and fine-tuning techniques reported a balanced performance \change{(BLEU score of 0.4049 for scene captioning and accuracy of 0.3508 for interaction detection) for both tasks on the target domain} and performed on-par with single-task models in domain adaptation.

\textbf{Conclusion:} 
The proposed multi-task model was able to adapt to domain shifts, incorporate novel instruments in the target domain, and perform tool-tissue interaction detection and report generation on par with single-task models. 

}

\keywords{Surgical scene understanding, Domain generalization, Scene graph, curriculum learning}

\maketitle

\section{Introduction}\label{Introduction}
Surgical scene understanding is of great significance to image-guided robot-assisted surgery~\cite{nwoye2020recognition,islam2020learning,qu2021development}. The computer-assisted robotic system integrated with scene understanding ability allows semi-automated or fully automated real-time supervision of robotic surgery and automated surgical report generation in the future. Given a surgical scene, inferring its surgical phase and related instrument-tissue interaction~\cite{islam2020learning,seenivasan2022global}, and automated caption generation~{\cite{xu2021learning,xu2021class} } will enable the development of intra-operative surgical supervision, postoperative analysis and report generation. However, recognizing the surgical phase and understanding the surgical activities in a complex surgery environment filled with blood, smoke, reflection and occlusion are challenging tasks. This task is further complicated when the model is presented with domain-shifted data. Theoretically, the model extracted features of a surgical tool used across various surgery should have the same feature descriptor. However, intensity shifts (tissue appearance, lighting and color contrast) between different surgeries cause the feature descriptor to differ. Inter-patient and intra-patient surgical domain shifts can also not be ignored. Furthermore, using surgery-specific instruments requires the model to be flexible to include new instruments. Varying surgical scenes with domain shifts and the use of surgery-specific instruments lead to the task of surgical domain adaptation (DA). A surgical scene understanding model equipped with the ability to deal with the DA can achieve better generalization and improve the surgical activity recognition outcome.  


\change{Works on scene graph for tool-tissue interaction detection and report generation has been recently reported. Graph-based models have been proposed to construct surgical scene graphs to detect tool-tissue interactions~\cite{islam2020learning,seenivasan2022global}. Works on recognizing surgical action triplets~\cite{nwoye2022rendezvous} have also been reported to detect tool-tissue interaction without the need to construct the scene graph. SGT~\cite{lin2022sgt} presented a scene graph-guided transformer to address the issue of surgical report generation. Richard et al.~\cite{bieck2021generation} proposed a keyword-augmented next sequence prediction approach that generates surgical reports based on keywords stated during panendoscopic procedures. While progress on independent tasks has been reported,} an automated surgical analysis system warrants multiple single-task learning (STL) models to output multiple parameters for integrated analysis. Compared to multiple STL models, a single multi-task learning \change{(MTL) model} can save computation, storage and maintenance costs~\cite{kokkinos2017ubernet,islam2020ap}. Additionally, MTL alleviates the problem of over-fitting in the model, improving the generalization of its shared sub-modules and utilizing the relationship between tasks to improve model prediction, enabling mutual assistance~\cite{islam2021st,seenivasan2022global}. However, training an MTL model is often challenging due to an imbalance in its task-specific performance convergence. To address DA in the surgical scene and reap the benefits of the MTL model, we propose an MTL model that provides competitive performance in both tool-tissue interaction detection and scene captioning tasks on both the source domain (SD) and the target domain (TD). To train the MTL model, we adopt the asynchronous task-aware optimization technique \cite{islam2020ap} that computes task-oriented gradients and trains the task-specific decoders independently. Our main contributions are summarized as follows:

\begin{itemize}
    \item Propose a novel multi-task model that can be domain adapted for surgical report generation and tool-tissue interaction.
    \item Address the domain adaptation in the surgical scene by combining the class incremental learning and supervised contrastive learning to handle the domain shift between the source and the target domain and novel instruments exist in the target domain.
    \item Develop the Laplacian of Gaussian based curriculum by smoothing  (CBS-LoG) to achieve better feature representation learning.
    \item Study task-aware optimization and distillation-based optimization technique improving convergence in MTL model.
\end{itemize}

\section{Methodology} \label{methodology}

\begin{figure*}[!t]
\centering
\includegraphics[width=1.0\linewidth]{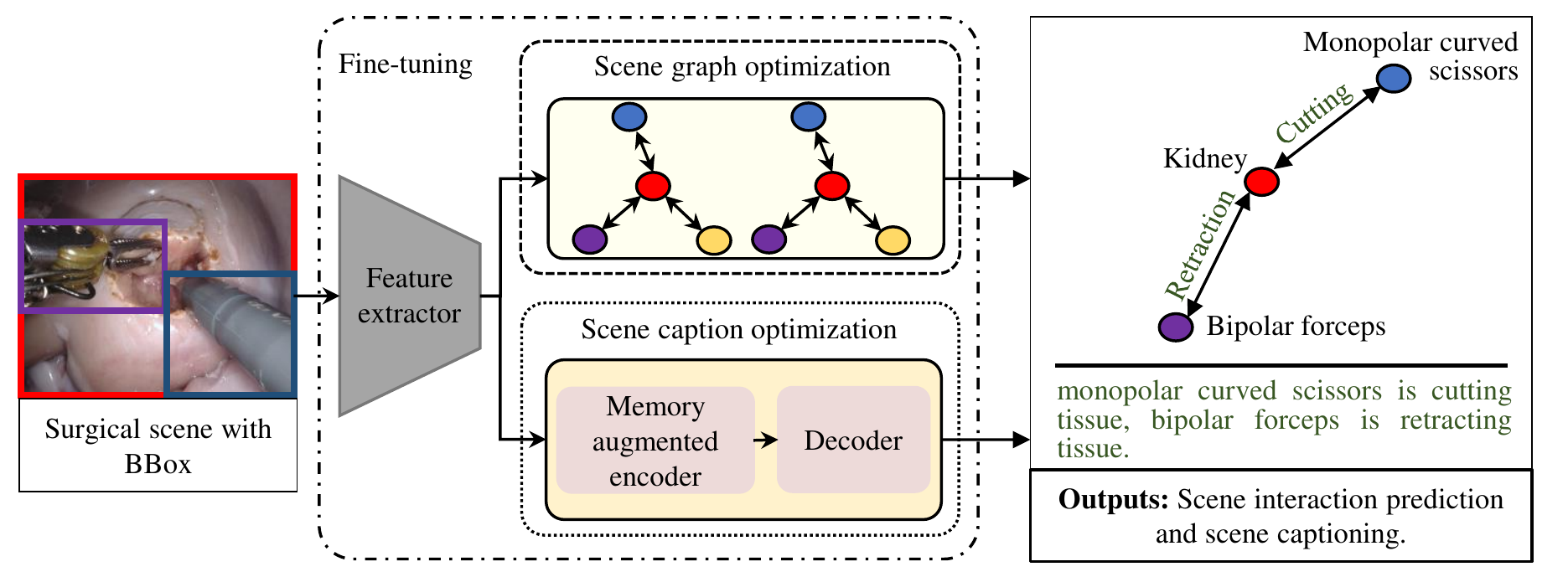}
\caption{Task-aware multi-task learning (MTL) model optimization and fine-tuning: The shared feature extractor is first initialized with weights trained for instrument classification. The MTL model is \change{first trained} by (i) freezing the feature extractor and scene graph and (ii) training the scene caption model until convergence. Secondly, (i) the feature extractor and scene graph weights are frozen and (ii) the scene graph model is trained until convergence. Finally, the complete MTL model is fine-tuned on both task losses.}
\label{fig:TA_MTL_optimization}
\end{figure*}

We propose an MTL model \change{(Fig.~\ref{fig:TA_MTL_optimization})} that performs tool-tissue interaction detection and scene captioning on both the \change{SD and TD}. The model consists of (i) a shared feature extractor, (ii) a transformer-based encoder-decoder network for scene captioning and a scene graph network for tool-tissue interaction \change{detection}. The shared feature extractor incorporates class incremental contrastive loss (CICL) to handle domain shifts. Furthermore, curriculum learning is applied to all three modules to enhance model learning. To train the MTL model, we adopt the asynchronous task-aware optimization technique \cite{islam2020ap} that computes task-oriented gradients and trains the task-specific \change{modules} independently. Vanilla and distillation-based optimization techniques are also explored \change{for MTL model training}.

\subsection{MTL Surgical Scene Understanding}

\subsubsection{Shared Feature Extractor}

A lightweight model ResNet18 \cite{he2016deep} is utilized as the feature extractor. To deal with the domain shift and the novel instruments that appear in the target domain, we design CICL by following\cite{xu2021class}. The contrastive learning loss minimizes the distance between the same label inputs across domains and pushes apart the samples with different labels in the feature embedding space. It can be formulated as:

\begin{equation}
    \mathcal{L}_{i}^{contra}=-\log \frac{\exp \left(\boldsymbol{z}_{i} \cdot \boldsymbol{z}_{j(i)} / \tau\right)}{\sum\nolimits_{k=1}^{2 N} \exp \left(\boldsymbol{z}_{i} \cdot \boldsymbol{z}_{k} / \tau\right)}
\end{equation}

\noindent where, temperature $\tau$, the logits of base sample $\boldsymbol{z}_{i}$, the logits of positive samples $\boldsymbol{z}_{j(i)}$ which have the same label with $\boldsymbol{z}_{i}$, the logits of negative samples $\boldsymbol{z}_{k}$ which have a different label with $\boldsymbol{z}_{i}$. When the loss is minimized, the numerator is expected to be higher and the denominator is lower, pushing everything else apart. The training method is designed to allow the model to learn incrementally, allowing new classes to be added. Similar to the KL-divergence mechanism in \cite{xu2021class}, the contrastive loss is used in updating the model weights for new classes. The total loss can be formulated as:

\begin{equation}
\mathcal{L}_{total} = \mathcal{L}_{contra} + \mathcal{L}_{incre}
\end{equation}

\subsubsection{Scene Caption}

\change{A} transformer-based encoder-decoder network \change{is built with inspiration from~\cite{cornia2020meshed} for scene captioning}. The network takes the \change{regions} of interest features from an \change{input image}, predicts the probability of words in the vocabulary, and generates the caption. The encoder consists of three encoder blocks which include the memory augmented self-attention layers. The decoder consists of three decoder blocks in which self-attention is applied to words and cross-attention is applied to all the encoder blocks outputs to understand the relationships among these \change{regions of interest}.

\subsubsection{Surgical Scene Graph}

Understanding the surgical scene and its interactions warrants a robust model that can adapt to the varying number of instruments and their interaction with the tissue. Inspired from VS-GAT \cite{liang2021visual}, we adopt the network for instrument-tissue interaction \change{detection}. Representing a scene as a graph ($\mathcal{G}$) with its nodes ($\mathcal{V}$) as the instruments and tissue, we use the graph edges ($\mathcal{E}$) for interaction classification. The scene graph model relies on the common feature extractor to extract features for the nodes. These (a) nodes features, together with (b) spatial features obtained based on bounding boxes and (c) semantic features obtained from word embedding \cite{mikolov2013distributed} are used by the graph model for interaction detection. While this model still needs to be adapted for domain shifts in visual features, the ability of the model to accommodate a varying number of instruments ($\mathcal{V}$) allows the model to incorporate novel instruments with ease.

\subsubsection{LoG-CBS}

Simulating a child in learning, curriculum by smoothing~\cite{sinha2020curriculum} employs Gaussian kernel to control the features entering the model at the initial epochs, allowing the model to learn gradually. The control over features passed to the model is achieved by controlling the $\sigma$. Decaying the $\sigma$ value every few epoch (epoch interval 5 by following~\cite{sinha2020curriculum} ) allow for more features to go through the model. Here, instead of the Gaussian kernel, we employ LOG kernels to control the features at the initial stages and highlight the instrument contours. LOG kernels are applied to all three sub-modules (i) feature extractor, (ii) scene graph and (iii) scene caption.

\begin{equation}
    LoG(x,y) = -{\frac{1}{\pi \sigma^4}}{\left[1-\frac{x^2+y^2}{2\sigma^2}\right]}{e^{-\frac{x^2+y^2}{2\sigma^2}}}
\end{equation}

\subsection{Model Optimization}

\subsubsection{MTL Optimization}

MTL models are often challenged by the asynchronous \change{performance} convergence of its independent \change{tasks}. As shown in \change{Fig.~\ref{fig:TA_MTL_optimization}}, we use task-aware MTL optimization to save the computation cost by sharing common networks and still achieve independent task model convergence, inspired from \cite{islam2020ap}. Here, the model optimization follows steps 1-4 in the algorithm~\ref{algortihm_ato}. The feature extractor is initially trained on instrument classification. The weights of the feature extractor and scene graph module are frozen. The scene caption \change{module} is then trained for scene captioning based on node features extracted using the feature extractor. Upon scene captioning task convergence, its weights are frozen. Finally, the scene graph \change{module} is trained to detect \change{tool}-tissue \change{interactions} based on the \change{node ($\mathcal{V}$) features} extracted using the feature extractor.

\begin{algorithm}[!h]
    \caption{\small{Task-Aware MTL Optimization and Fine-Tuning}} \label{algortihm_ato}
    \begin{algorithmic}[1]
        \small
        \State \textbf{Initialize model weights} \\
            Shared (${W_{sh}}$), scene graph (${W_{sg}}$),  caption (${W_c}$) \\
        
        \State \textbf{[Set gradient accumulators to zero]} \\
            shared (${dW_{sh}}$), scene graph (${dW_{sg}}$),  caption (${dW_c}$) \\
            $\mathbf{dW_{sh}} \leftarrow 0,\; \mathbf{dW_{d}} \leftarrow 0,\; \mathbf{dW_{s}} \leftarrow 0$\\
        
        \State \textbf{[Optimize Caption Network]}
        \While {$caption\; task\; not\; converged\;$}
        \State \textit{[caption gradients w.r.t caption loss ${L_{s}}$]}
        \State $\mathbf{dW_{c}} \leftarrow \mathbf{dW_{c}} + \sum_{i}\delta_{i}\nabla_{W_{c}}L_{c}(W_{sh}, W_{c})$
        \EndWhile \\
        
        \State \textbf{[Optimize Scene Graph]}
        \While {$scene\; graph\; task\; not\; converged\;$}
        \State \textit{[Scene graph block gradients w.r.t scene graph loss ${L_{sg}}$]}
        \State $\mathbf{dW_{sg}} \leftarrow \mathbf{dW_{sg}} + \sum_{i}\delta_{i}\nabla_{W_{sg}}L_{sg}(W_{sh},W_{sg})$
        \EndWhile \\

        \State \textbf{[Fine-tuning]}
        \While{$both\; tasks\; improving\;$}
        \State $\mathbf{dW_{sh}} \leftarrow \mathbf{dW_{sh}} + \sum_{i}\delta_{i}\nabla_{W_{sh}}L(W_{sh},W_{sg},W_{c})$
        \State $\mathbf{dW_{sg}} \leftarrow \mathbf{dW_{sg}} + \sum_{i}\delta_{i}\nabla_{W_{sg}}L(W_{sh},W_{sg},W_{c})$
        \State $\mathbf{dW_{c}} \leftarrow \mathbf{dW_{c}} + \sum_{i}\delta_{i}\nabla_{W_{c}}L(W_{sh},W_{sg},W_{c})$
        \EndWhile
    \end{algorithmic}
\end{algorithm}

\subsubsection{MTL Optimization and Fine-Tuning (MTL-FT)}
The task-aware MTL optimization ensures that the convergence of the independent task module’s performance is optimal and on-par with single-task model performance. However, this forfeits the model generalization advantage of MTL models. Here, this drawback is addressed by further fine-tuning the task-aware MTL model on combined task loss (Algorithm \ref{algortihm_ato} step: 1-5). The combined loss is given by:

\begin{equation}
    \mathcal{L}_{fine\_tune} = 0.5 * (\mathcal{L}_{caption} +  \mathcal{L}_{scene\_graph})
\end{equation}

\subsubsection{MTL Vanilla (MTL-V) optimization}
Under MTL-V optimization, the feature extractor is initially loaded with pre-trained classification weights. The complete MTL model is then fully trained based on the scene caption’s and the scene graph’s combined loss.
 
 \begin{equation}
    \mathcal{L}_{MTL\_Vanilla} = 0.5 * (\mathcal{L}_{caption} +  \mathcal{L}_{scene\_graph})
\end{equation}

\begin{algorithm}[!h]
    \caption{\small{Knowledge distillation-based MTL Optimization}} \label{algortihm_mtl_kd}
    \begin{algorithmic}[1]
        \small
        \State \textbf{[Initialize model weights]} \\
        shared (${W_{sh}}$), scene graph (${W_{sg}}$),  caption (${W_c}$)\\
        
        \State \textbf{[Training]}
        \While{$both\; tasks\; improving\;$}
            \State $\mathbf{dW_{sh}} \leftarrow \mathbf{dW_{sh}} + \sum_{i}\delta_{i}\nabla_{W_{sh}}L(W_{sh},W_{sg},W_{c})$
            \State $\mathbf{dW_{sg}} \leftarrow \mathbf{dW_{sg}} + \sum_{i}\delta_{i}\nabla_{W_{sg}}L(W_{sh},W_{sg},W_{c})$
            \State $\mathbf{dW_{c}} \leftarrow \mathbf{dW_{c}} + \sum_{i}\delta_{i}\nabla_{W_{c}}L(W_{sh},W_{sg},W_{c})$
        \EndWhile
        
    \end{algorithmic}
\end{algorithm}

\subsubsection{MTL \change{Knowledge} Distillation (MTL-\change{K}D)}

\begin{figure}[!h]
\centering
\includegraphics[width=0.8\linewidth]{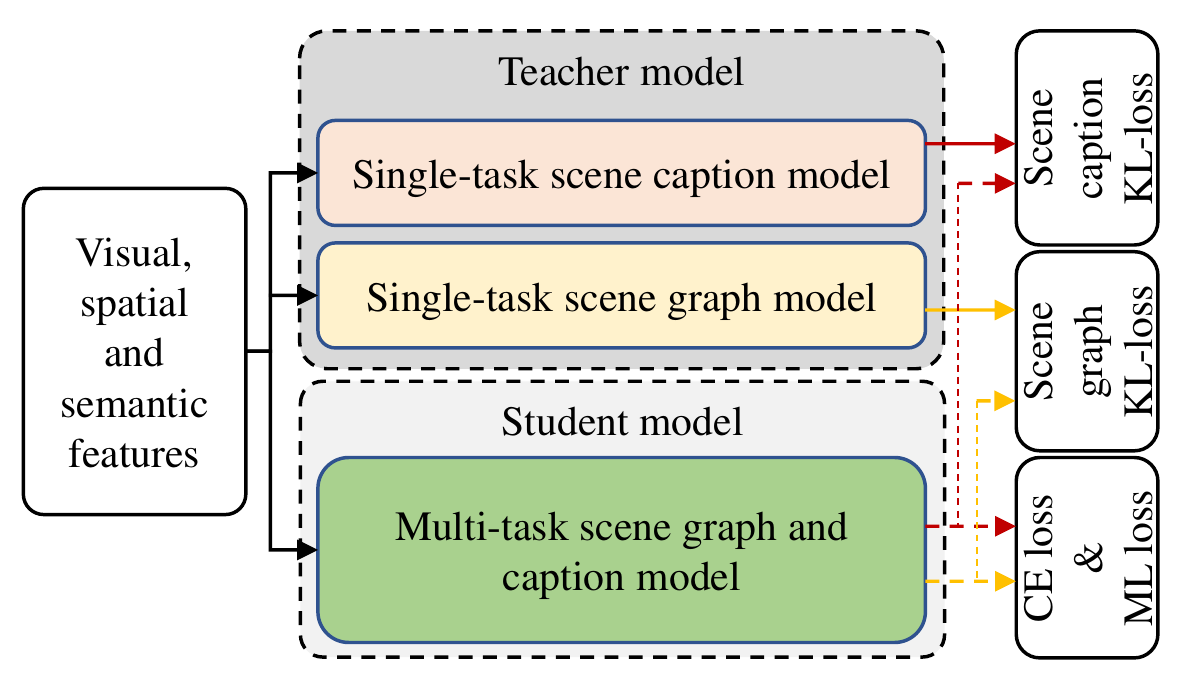}
\caption{MTL-Knowledge Distillation: The MTL model is trained based on both task prediction vs the grouth truth loss and Kullback-Leibler divergence loss between the MTL model and single-task models (Teacher model).}
\label{fig:MTL_distillation}
\end{figure}

\change{A} knowledge distillation-based MTL \change{(MTL-KD)} optimization technique (Fig.~\ref{fig:MTL_distillation}), is also explored. This technique is adopted to reduce the effect of asynchronous model convergence. During training, the MTL model is trained based on both (i) the prediction vs. the ground truth loss and (ii) Kullback-Leibler(KL) divergence loss between the MTL model’s and STL model’s logits (Algorithm \ref{algortihm_mtl_kd}). The MTL-\change{K}D loss ($\mathcal{L}_{MTL-\change{K}D} $) function is given by:
\begin{equation}
    \begin{split}
    \mathcal{L}_{MTL-\change{K}D} =  0.35 * (\mathcal{L}_{CE\_c} +  \mathcal{L}_{ MLS\_sg}) 
    + 0.15 * (\mathcal{L}_{KL\_c} +  \mathcal{L}_{KL\_sg})
    \end{split}
\end{equation}
where, $\mathcal{L}_{CE\_c}$ is the scene caption model’s cross-entropy loss. $\mathcal{L}_{ MLS\_sg}$ is the scene graph model’s multi-label loss, $\mathcal{L}_{KL\_c}$ is the scene caption model’s KL divergence loss and $\mathcal{L}_{KL\_sg}$ is the scene graph model’s KL-divergence loss.

\section{Experiment} \label{experiment}

\subsection{Dataset} \label{dataset}

\subsubsection{Source Domain}
The \change{SD} dataset comes from the training set of \change{the} MICCAI robotic instrument segmentation dataset of \change{the} endoscopic vision challenge 2018 \cite{allan20202018}. The dataset contains $8$ instruments \change{and} $1$ organ (kidney). \change{$11$ tool-tissue interactions are used to generate the captions for the scenes.} The interaction annotation is from \cite{islam2020learning} and the caption annotation is from \cite{xu2021class}. \change{The dataset is split into two:} the validation subset with $447$ labelled frames ($1^{st}$, $5^{th}$ \change{and} $16^{th}$ sequences) and the training subset with $1560$ labelled frames (the remaining $11$ sequences) after removing some frames with no interaction by following the work~\cite{islam2020learning, xu2021learning, xu2021class}.


\subsubsection{Target Domain}
The \change{TD} dataset is from 23 surgical videos, which contain 22 transoral robotic surgery operations provided by the hospitals and 1 robotic nephroureterectomy video\footnote{https://youtu.be/bwpEul4KCSc} considering the need to balance the dataset. \change{The TD dataset consists of 4 instruments (two novel instruments) and 1 tissue (Oropharynx).} The interaction annotation and caption annotation are from \cite{xu2021class}, \change{based on the $5$ kinds of interactions.} The training subset includes sequences 1-13th, 20th \change{and} 23rd (77 labelled frames), and the validation subset includes the remaining sequences (258 labelled frames). 


\change{Additional details on these two datasets are provided in the \textbf{supplementary file}.}

\subsection{Implementation details} \label{implementation_details}
The MTL model and its variants are implemented using the PyTorch framework\footnote{https://github.com/lalithjets/Domain-adaptation-in-MTL}. \change{Firstly}, the feature extractor is trained on instrument classification using the incremental learning technique and the \change{supervised} contrastive loss. The MTL model is then trained using different optimization techniques by \change{employing} cross-entropy loss for caption task and multi-label loss for interaction detection. For optimization variants that perform task-aware training. In the MTL combined task fine-tuning stage and other MTL optimization variants that train both tasks simultaneously. \change{All training hyperparameters are provided in the \textbf{supplementary file}}. For unsupervised domain \change{adaptation} (UDA), The model is trained only on the SD, with its feature extractor loaded with weights trained on classifying instruments present in the SD. For DA, the model is initially trained on SD and domain adapted to TD using few-shot learning. Here, the feature extractor is loaded with weights trained using incremental learning to include new instruments in the TD domain. All SOTA benchmarked models\footnote{https://github.com/birlrobotics/vs-gats}\footnote{https://github.com/aimagelab/meshed-memory-transformer} in the work are re-implemented using its author’s official Github code and re-trained on our datasets.

\section{Results} \label{results}

\begin{table*}[!h]
\centering
\caption{Comparison of our proposed domain adapted MTL model and its variants against state-of-the-art models for tool-tissue-interaction detection and scene captioning in source and target domain. UDA and Few refer to unsupervised domain adaptation and few-shot learning techniques, respectively.}
\scalebox{0.5}{
\change{
\begin{tabular}{c|c|c|c|c|c|c|c|c|c|c|c|c}
\toprule
\multirow{3}{*}{\textbf{Model}}                                      & \multicolumn{2}{c|}{\multirow{3}{*}{\textbf{Regime}}}                                                  & \multicolumn{4}{c|}{\textbf{Caption}}                                                               & \multicolumn{6}{c}{\textbf{Interaction}}                                                                                             \\ \cline{4-13} 
                                                            & \multicolumn{2}{c|}{}                                                                         & \multicolumn{2}{c|}{\textbf{SD}}                 & \multicolumn{2}{c|}{\textbf{TD}}                                   & \multicolumn{3}{c|}{\textbf{SD}}                                      & \multicolumn{3}{c}{\textbf{TD}}                                       \\ \cline{4-13} 
                                                            & \multicolumn{2}{c|}{}                                                                         & \textbf{BLEU}               & \textbf{CIDEr}              & \textbf{BLEU}               & \textbf{CIDEr}              & \textbf{Acc}                & \textbf{mAP}                & \textbf{Recall}             & \textbf{Acc}                & \textbf{mAP}                & \textbf{Recall}             \\
\toprule
\multirow{2}{*}{M2T\cite{cornia2020meshed}}                 & \multirow{2}{*}{Caption}                                      & UDA                           & 0.5703             & 2.5385             & 0.2897             & 0.2511             & \multirow{2}{*}{-} & \multirow{2}{*}{-} & \multirow{2}{*}{-} & \multirow{2}{*}{-} & \multirow{2}{*}{-} & \multirow{2}{*}{-} \\ \cline{3-7}
                             &                                                                                              & Few                            & 0.331              & 0.1825             & 0.6066             & 3.1188             &                    &                    &                    &                    &                    &                    \\ \hline
\multirow{2}{*}{Xu et al. \cite{xu2021learning}}      & \multirow{2}{*}{Caption}                                            & UDA                           & 0.5875             & 2.5930             & 0.2100             & 0.2029             & \multirow{2}{*}{-} & \multirow{2}{*}{-} & \multirow{2}{*}{-} & \multirow{2}{*}{-} & \multirow{2}{*}{-} & \multirow{2}{*}{-} \\ \cline{3-7}
                             &                                                                                              & Few                            & 0.2645             & 0.2029             & 0.5768             & 2.8619             &                    &                    &                    &                    &                    &                    \\ \hline
\multirow{2}{*}{GPNN~\cite{qi2018learning}}        & \multirow{2}{*}{Interaction}                                           & UDA                           & \multirow{2}{*}{-} & \multirow{2}{*}{-} & \multirow{2}{*}{-} & \multirow{2}{*}{-} & 0.4957             & 0.2033             & -                  & 0.3057             & 0.2031             & -                  \\ \cline{3-3} \cline{8-13} 
                             &                                                                                              & Few                            &                    &                    &                    &                    & 0.2916             & 0.2426             & -                  & 0.3399             & 0.2271             & -                  \\ \hline
\multirow{2}{*}{Islam et al.~\cite{islam2020learning}}    & \multirow{2}{*}{Interaction}                                    & UDA                           & \multirow{2}{*}{-} & \multirow{2}{*}{-} & \multirow{2}{*}{-} & \multirow{2}{*}{-} & 0.4802             & 0.2157             & -                  & 0.3144             & 0.1996             & -                  \\ \cline{3-3} \cline{8-13} 
                             &                                                                                              & Few                            &                    &                    &                    &                    & 0.2765             & 0.1913             & -                  & 0.3975             & 0.2086             & -                  \\ \hline
\multirow{2}{*}{VS-GAT~\cite{liang2021visual} }      & \multirow{2}{*}{Interaction}                                         & UDA                           & \multirow{2}{*}{-} & \multirow{2}{*}{-} & \multirow{2}{*}{-} & \multirow{2}{*}{-} & 0.6305             & 0.2658             & 0.2868                  & 0.3319             & 0.0777             & 0.0949                  \\ \cline{3-3} \cline{8-13} 
                             &                                                                                              & Few                            &                    &                    &                    &                    & 0.3867             & 0.2219             &    0.2083            & 0.3377             & 0.0812             & 0.1097                  \\ \hline
\multirow{4}{*}{Ours}       & \multirow{2}{*}{\begin{tabular}[c]{@{}c@{}}MTL\end{tabular}}                              & UDA                           & 0.5                & 2.0261             & 0.2685             & 0.2043             & 0.6434             & \textbf{0.3041}             & 0.2956             & 0.3479             & 0.0789             & 0.0975             \\ \cline{3-13} 
                             &                                                                                              & Few                           & 0.2204             & 0.1146             & \textbf{0.5232}    & \textbf{2.5304}    & 0.4022             & 0.2269             & 0.2198             & 0.3479             & 0.0920             & 0.1180             \\ \cline{2-13} 
                             & \multirow{2}{*}{\begin{tabular}[c]{@{}c@{}}MTL-FT \end{tabular}}                          & UDA                      & \textbf{0.6049}             & \textbf{2.4098}             & 0.1758             & 0.2084             & \textbf{0.6606}             & 0.2951             & \textbf{0.3033}    & 0.3290             & 0.0753             & 0.0826             \\ \cline{3-13} 
                             &                                                                                              & Few                      & 0.2531             & 0.0767             & 0.4049             & 2.4075             & 0.3781             & 0.2120             & 0.2206             & \textbf{0.3508}             & \textbf{0.0968}             & \textbf{0.1236}    \\ 
\bottomrule
\end{tabular}
}
}
\label{SOTA_performance}
\end{table*}

\begin{figure*}[!t]
\centering
\includegraphics[width=0.9\linewidth]{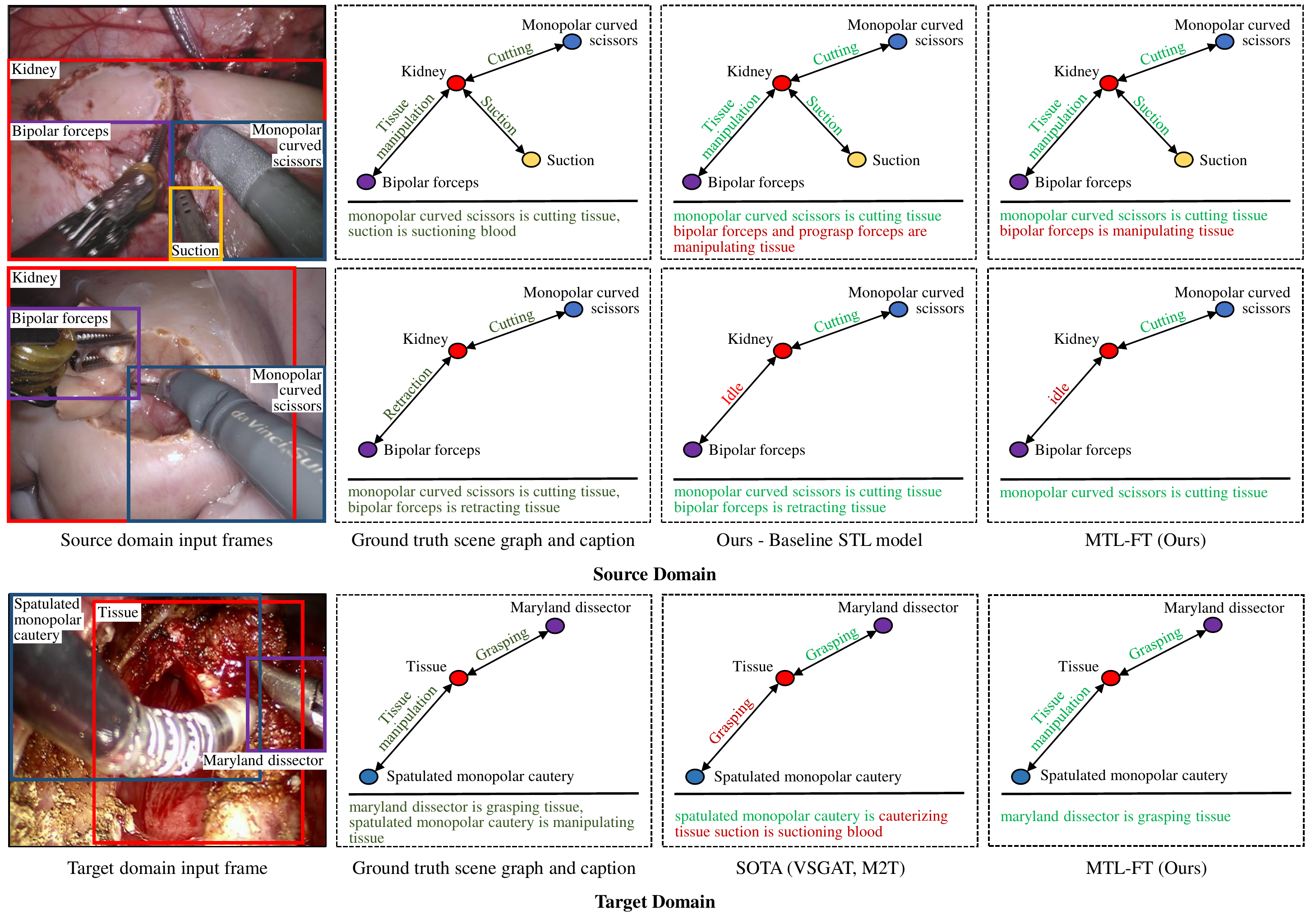}
\caption{Qualitative analysis: Comparison of our proposed task-aware optimized and fine-tuned multi-task model performance vs single-task models performance and GT in detecting tool-tissue interaction and surgical scene captioning.}
\label{fig: Qualitative_results}
\end{figure*}

The captioning and interaction detection performance of the domain-adapted MTL model on both SD and TD is reported quantitatively and qualitatively. The model’s performance in captioning task is quantified using the BLEU\cite{papineni2002bleu} and CIDEr\cite{vedantam2015cider} scores. The interaction detection task performance is quantified using accuracy (Acc), mean average precision (mAP) and Recall. From Table. \ref{SOTA_performance}, is it observed that in most cases, MTL models in their respective best task performances outperform SOTA models. Within MTL models, the best performance in caption tasks is split between \change{MTL-FT (task-aware optimization and fine-tuning)} for SD and task-aware MTL for TD. \change{However, the best performance among MTL models on both SD and TD for interaction detection is observed in MTL-FT.} While a diverse performance in MTL models is expected due to asynchronous convergence in independent task performance, a balanced performance on both tasks is observed in the MTL-FT (task-aware MTL optimization and fine-tuning) model. Qualitative analysis (Fig. \ref{fig: Qualitative_results}) comparing the STL model and MTL-FT also shows that MTL-FT performs on par with STL models in both SD and TD.

\begin{table*}[!h]
\centering
\caption{Ablation study on the performance of the proposed multi-task model when trained using (i) task-aware optimization and fine-tuning, (ii) vanilla optimization, (iii) distillation based optimization and (iv) distillation based optimization and finetuning. BG and BC refer to the best graph and best caption weights.}
\scalebox{0.54}{
\begin{tabular}{c|c|c|c|c|c|c|c|c|c|c|c}
\toprule
\multicolumn{2}{c|}{\multirow{3}{*}{\textbf{Regime}}} & \multicolumn{4}{c|}{\textbf{Caption}}                                 & \multicolumn{6}{c}{\textbf{Interaction}}                                                                                      \\ \cline{3-12} 
\multicolumn{2}{c|}{}                                 & \textbf{SD}     & \textbf{}       & \multicolumn{2}{c|}{\textbf{TD}}  & \multicolumn{3}{c|}{\textbf{SD}}                    & \multicolumn{3}{c}{\textbf{TD}}                                         \\ \cline{3-12} 
\multicolumn{2}{c|}{}                                 & \textbf{BLEU}   & \textbf{CIDEr}  & \textbf{BLEU}   & \textbf{CIDEr}  & \textbf{Acc}    & \textbf{mAP}    & \textbf{RECALL} & \textbf{Acc}    & \textbf{mAP}    & \multicolumn{1}{c}{\textbf{RECALL}} \\
\toprule
\multirow{4}{*}{\textbf{MTL-FT}}          & UDA-BG   & 0.6049          & 2.4098          & 0.1758          & 0.2084          & 0.6606          & 0.2951          & \textbf{0.3033} & 0.3290          & 0.0753          & 0.0826                               \\ \cline{2-12} 
                                         & UDA-BC   & 0.5652          & 2.5902          & 0.1715          & 0.2305          & 0.6400          & 0.3052          & 0.2991          & 0.3246          & 0.073           & 0.0852                               \\ \cline{2-12} 
                                         & Few-BG   & 0.2128          & 0.0703          & 0.3446          & 2.1511          & 0.3859          & 0.2162          & 0.2226          & 0.3624          & \textbf{0.0974} & 0.1227                               \\ \cline{2-12} 
                                         & Few-BC   & 0.2531          & 0.0767          & \textbf{0.4049} & \textbf{2.4075} & 0.3781          & 0.2120          & 0.2206          & 0.3508          & 0.0968          & \textbf{0.1236}                      \\ \hline
\multirow{4}{*}{\textbf{MTL-V}}    & UDA-BG   & 0.5502          & 2.2516          & 0.1375          & 0.1938          & \textbf{0.6684} & \textbf{0.3132} & 0.2974          & 0.3071          & 0.0768          & 0.0930                               \\ \cline{2-12} 
                                         & UDA-BC   & \textbf{0.6167} & \textbf{3.0918} & 0.1052          & 0.1189          & 0.5866          & 0.2723          & 0.2173          & 0.3828          & 0.0802          & 0.1019                               \\ \cline{2-12} 
                                         & Few-BG   & 0.4292          & 1.1167          & 0.239           & 0.4769          & 0.5237          & 0.2400          & 0.1993          & 0.3508          & 0.0718          & 0.0857                               \\ \cline{2-12} 
                                         & Few-BC   & 0.2634          & 0.5294          & 0.3236          & 1.2506          & 0.3411          & 0.1939          & 0.2096          & 0.2766          & 0.0765          & 0.0933                               \\ \hline
\multirow{4}{*}{\textbf{MTL-KD}}         & UDA-BG   & 0.5429          & 1.9870          & 0.1434          & 0.1531          & 0.6288          & 0.2119          & 0.2178          & 0.377           & 0.0862          & 0.1090                               \\ \cline{2-12} 
                                         & UDA-BC   & 0.5765          & 2.6308          & 0.1452          & 0.2963          & 0.6012          & 0.2311          & 0.2039          & 0.2955          & 0.0832          & 0.0909                               \\ \cline{2-12} 
                                         & Few-BG   & 0.5848          & 2.7290          & 0.3227          & 0.7236          & 0.5874          & 0.234           & 0.2011          & \textbf{0.4003} & 0.0839          & 0.1085                               \\ \cline{2-12} 
                                         & Few-BC   & 0.2884          & 0.0783          & 0.2839          & 1.6126          & 0.4496          & 0.2091          & 0.1189          & 0.3479          & 0.0856          & 0.1042                               \\ \hline
\multirow{4}{*}{\textbf{MTL-KD-FT}}      & UDA-BG   & 0.5605          & 2.0041          & 0.2395          & 0.1302          & 0.6658          & 0.2934          & 0.2860          & 0.2722          & 0.0812          & 0.0794                               \\ \cline{2-12} 
                                         & UDA-BC   & 0.5455          & 2.2414          & 0.1887          & 0.1087          & 0.5401          & 0.2382          & 0.1948          & 0.3246          & 0.0804          & 0.0907                               \\ \cline{2-12} 
                                         & Few-BG   & 0.3276          & 0.1705          & 0.3429          & 1.2957          & 0.4617          & 0.1962          & 0.1237          & 0.3668          & 0.0912          & 0.1127                               \\ \cline{2-12} 
                                         & Few-BC   & 0.3457          & 0.2059          & 0.3093          & 1.5961          & 0.4539          & 0.2074          & 0.1473          & 0.3071          & 0.0869          & 0.1006                               \\ \bottomrule
\end{tabular}}
\label{ablation}
\end{table*}

\subsection{Ablation Study}
Optimization techniques strongly influence the MTL model's training due to their asynchronous performance convergence. In Table~\ref{ablation}, we also report an ablation study on performance between (i) MTL-FT, (ii) MTL-V, (iii) MTL-KD and (iv) MTL-KD-FT models. While distillation-based MTL training techniques have been proposed to improve MTL performance,  based on Table~\ref{ablation}, it is observed that it’s on a case-by-case basis. In our model, MTL-FT and MTL-V have outperformed distillation-based MTL models in most cases. 

\section{Discussion}
\change{Although our MTL-FT achieved a balanced performance on both captioning and interaction detection tasks in UDA and Few-shot DA (Table~\ref{SOTA_performance}), as observed in Table~\ref{ablation}, achieving optimal convergence (best performance) of both tasks remains a challenge to address. Compared between UDA and Few-Shot DA in Table~\ref{SOTA_performance}, UDA performance is better on SD compared to Few-shot which performs better on TD. This is an expected result as the model is further trained only on TD in a few-shot DA. As this work is trained and validated on limited datasets, further validations on additional datasets can be done as future work to test for any bias.}

\section{Conclusion} \label{Conclusion}
This paper proposes an MTL model that can be adapted to domain shift in surgical scenes for surgical report generation and tool-tissue interaction detection. To address the intensity shift and include novel instruments in the new domain, we incorporate supervised contrastive learning and class incremental learning into the feature extractor. We further introduce Laplacian of Gaussian kernel-based curriculum learning into the MTL model’s shared and task-specific modules, allowing the model to learn better. The performance of the MTL model trained based on task-aware optimization and fine-tuning is observed to per on-par with SOTA models. This results in MTL surgical scene understanding model that can adapt to domain shift in performing instrument-tissue interaction detection and scene captioning. On top of employing task-aware optimization techniques, we also study vanilla optimization and distillation-based MTL optimization and conclude that task-aware optimization is favorable for balanced performance between the tasks.

\section{List of Abbreviations}
\begin{itemize}
    \item[--] \change{Acc: Accuracy}
    \item[--] \change{CBS: Curriculum By Smoothing}
    \item[--] \change{CICL: Class Incremental Contrastive Learning}
    \item[--] \change{DA: Domain Adaptation}
    \item[--] \change{LoG: Laplacian of Gaussian}
    \item[--] \change{mAP: Mean Average Precision}
    \item[--] \change{MTL : Multi-Task Learning}
    \item[--] \change{MTL-FT: Task-Aware Multi-Task Learning Optimization and Fine-Tuning}
    \item[--] \change{MTL-KD: MTL based on Knowledge Distillation}
    \item[--] \change{MTL-KD-FT: MTL based on Knowledge Distillation and Fine-Tuning}
    \item[--] \change{MTL-V: MTL Vanilla}
    \item[--] \change{SD: Source Domain}
    \item[--] \change{SOTA: State-Of-The-Art}
    \item[--] \change{STL: Single-Task Learning}
    \item[--] \change{TD: Target domain}
    \item[--] \change{UDA: Unsupervised Domain Adaptation}
\end{itemize}

\backmatter

\section{Declarations}

\subsection{Code availability}
This work is implemented using the PyTorch framework and the codes are available at: \href{https://github.com/lalithjets/Domain-adaptation-in-MTL}{https://github.com/lalithjets/Domain-adaptation-in-MTL}.

\subsection{Funding}
This work was supported by the National Key R\&D Program of China under Grant 2018YFB1307700 (with subprogram 2018YFB1307703) from the Ministry of Science and Technology (MOST) of China, the Shun Hing Institute of Advanced Engineering (SHIAE project BME-p1-21, 8115064) at the Chinese University of Hong Kong (CUHK), and Singapore Academic Research Fund under Grant R397000353114.

\subsection{Ethics approval}
All procedures performed in studies involving human
participants were in accordance with the ethical standards of the institutional and/or national research committee.

\subsection{Consent to participate}
This article does not contain patient data.

\subsection{Conflict of interest} 
The authors declare that they have no conflict of interest.

\bibliography{sn-bibliography}

\end{document}